\documentclass[10pt,twocolumn,letterpaper]{article}

\usepackage{cvpr}              

\usepackage{graphicx}
\usepackage{amsmath}
\usepackage{amssymb}
\usepackage{booktabs}
\usepackage{multirow}
\usepackage{epstopdf}
\usepackage{pifont}
\usepackage{appendix}
\usepackage[misc]{ifsym}

\usepackage[pagebackref,breaklinks,colorlinks]{hyperref}

\usepackage[capitalize]{cleveref}
\crefname{section}{Sec.}{Secs.}
\Crefname{section}{Section}{Sections}
\Crefname{table}{Table}{Tables}
\crefname{table}{Tab.}{Tabs.}

\def\cvprPaperID{****} 
\def\confName{CVPR}
\def\confYear{2023}

\begin{document}




\title{Texts as Images in Prompt Tuning for Multi-Label Image Recognition}

\author{
  {\small Zixian Guo\textsuperscript{\rm 1}\thanks{This work was done when Zixian Guo was a research intern at TAL.} \quad Bowen Dong\textsuperscript{\rm 2}\quad Zhilong Ji\textsuperscript{\rm 1}\quad Jinfeng Bai\textsuperscript{\rm 1}\quad Yiwen Guo\textsuperscript{\rm 4}\quad  Wangmeng Zuo\textsuperscript{\rm 2,3}$^{\textrm{\Letter}}$} \\
  {\small \textsuperscript{\rm 1}Tomorrow Advancing Life \quad \textsuperscript{\rm 2}Harbin Institute of Technology \quad \textsuperscript{\rm 3}Pazhou Lab, Guangzhou \quad \textsuperscript{\rm 4}Independent Researcher}\\
  \tt{\small{zixian\_guo@foxmail.com \quad cndongsky@gmail.com \quad zhilongji@hotmail.com}} \\
  \tt{\small{jfbai.bit@gmail.com \quad guoyiwen89@gmail.com \quad wmzuo@hit.edu.cn}}
}

\maketitle

\begin{abstract}
    Prompt tuning has been employed as an efficient way to adapt large vision-language pre-trained models (\eg CLIP) to various downstream tasks in data-limited or label-limited settings. 
    Nonetheless, visual data (\eg, images) is by default prerequisite for learning prompts in existing methods.
    In this work, we advocate that the effectiveness of image-text contrastive learning in aligning the two modalities (for training CLIP) further makes it feasible to treat texts as images for prompt tuning and introduce TaI prompting. 
    In contrast to the visual data, text descriptions are easy to collect, and their class labels can be directly derived.
    Particularly, we apply TaI prompting to multi-label image recognition, where sentences in the wild serve as alternatives to images for prompt tuning.
    Moreover, with TaI, double-grained prompt tuning (TaI-DPT) is further presented to extract both coarse-grained and fine-grained embeddings for enhancing the multi-label recognition performance.
    Experimental results show that our proposed TaI-DPT outperforms zero-shot CLIP by a large margin on multiple benchmarks, \eg, MS-COCO, VOC2007, and NUS-WIDE, while it can be combined with existing methods of prompting from images to improve recognition performance further.
    Code is released at https://github.com/guozix/TaI-DPT.
   
\end{abstract}

\begin{figure}
  \centering
     \includegraphics[width=0.95\linewidth]{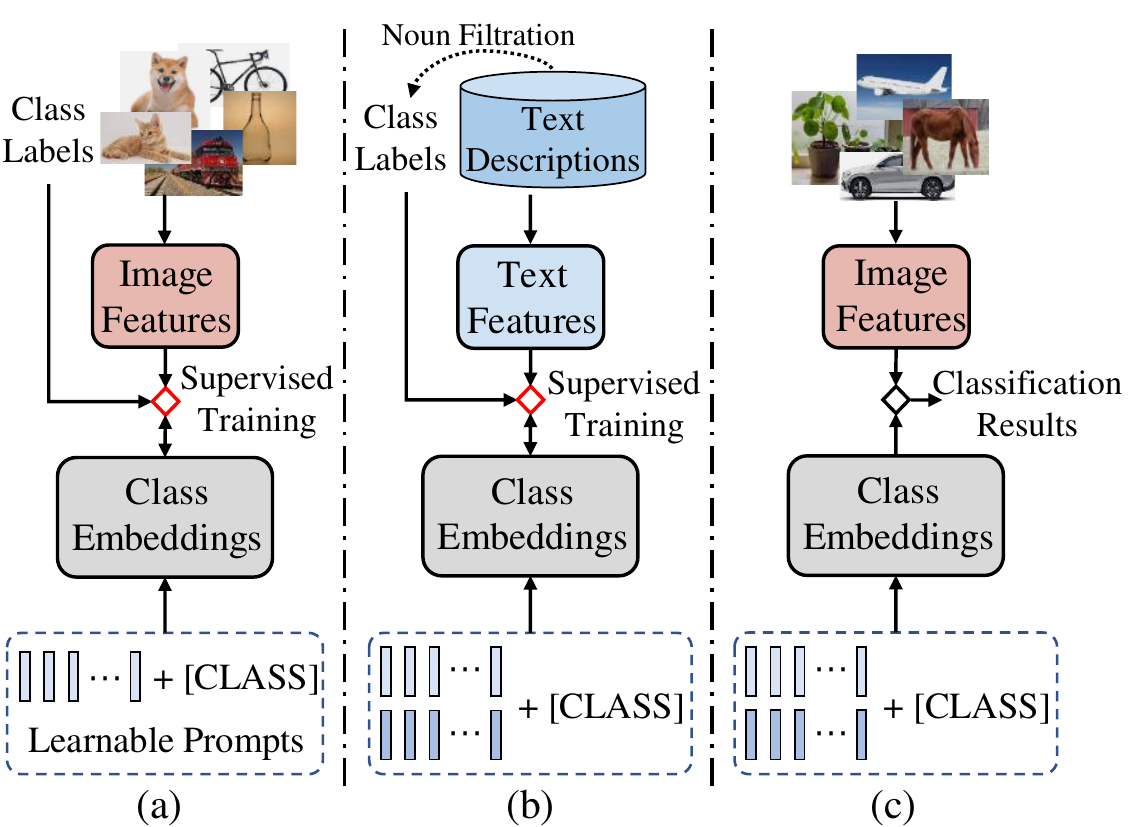}
  \vspace{-2mm}
  \caption{A comparison between prompting from images and our text-as-image (TaI) prompting. (a) Prompting from images (\eg,~~\cite{coop}) uses labeled images of task categories to learn the text prompts. Instead, (b) our TaI prompting learn the prompts with easily-accessed text descriptions containing target categories. (c) After training, the learned prompts in (a) or (b) can be readily applied to test images.}
  \label{fig:intro}
\end{figure}

\section{Introduction}
\label{sec:introduction}
Recent few years have witnessed rapid progress in large vision-language (VL) pre-trained models~\cite{clip,align,declip,filip,alayrac2022flamingo,yuan2021florence} as well as their remarkable performance on downstream vision tasks.
A VL pre-trained model generally involves data encoders and it is becoming increasingly popular to exploit image-test contrastive loss~\cite{clip} to align the embedding of images and texts into a shared space.  
When adapting to downstream tasks in relatively data-limited or label-limited settings, it is often ineffective to fine-tune the entire model, due to its high complexity.
Then, prompt tuning as a representative parameter-efficient learning paradigm has emerged as an efficient way to adapt VL model to downstream tasks.



Albeit considerable achievements have been made, existing prompt tuning methods generally require visual data to learn prompts (as shown in Fig.~\ref{fig:intro}(a)). 
For example, CoOp~\cite{coop} learns from annotated images.
CoCoOp~\cite{cocoop} further introduces generalizable input-conditional prompts.
DualCoOp~\cite{dualcoop} adapts CLIP to multi-label recognition tasks by training pairs of positive and negative prompts with partial-labeled images.
Nonetheless, the performance of these prompting methods may be limited when it is infeasible to obtain sufficient image data or annotate the required images.


In this paper, we advocate treating \textbf{T}exts \textbf{a}s \textbf{I}mages for prompt tuning, \ie, TaI prompting. 
It is considered feasible as the image encoder and text encoder in many pre-trained VL models~\cite{clip,align} encode images and texts into a shared space. 
Given an image and its caption, the visual features produced by the image encoder will be close to the text feature of the caption produced by the text encoder.  
Therefore, in addition to extracting visual features from images, it is also feasible to extract text features as alternatives form, for example, descriptive sentences and captions, for prompt tuning (see Fig.~\ref{fig:intro}(b)). 
TaI prompting has several interesting properties and merits.
Taking a downstream image recognition task as an example, given a set of object categories, one can easily crawl a large set of text descriptions that contain object names from these categories. 
Text descriptions are easily accessible in this way, and class labels can be directly derived from text descriptions, which means, 
in contrast to prompting from images, TaI prompting may suffer less from the data-limited and label-limited issues.

We use multi-label image recognition~\cite{voc2007,coco,nuswide,chen2019multi,ye2020add} to verify the effectiveness of our TaI prompting in this paper. 
To begin with, we crawl the captions from public image caption datasets (\eg, MS-COCO~\cite{coco}) and localized narratives from object detection datasets (\eg, Open
Images~\cite{openimages}) to form the training set of text descriptions.
For any specific multi-label recognition task, we adopt a noun filter to map the nouns in the text descriptions to the corresponding object categories, and then only keep the text descriptions that contain one or more classes of target objects.
To better cope with multi-label classification, we introduce double-grained prompt tuning (\ie, TaI-DPT) which involves: (i) a set of global prompts to generate embeddings for classifying whole sentences or images, and (ii) a set of local prompts to extract embeddings for discriminating text tokens or image patches. 
Given a set of text descriptions, global and local prompts can be tuned by minimizing the ranking loss~\cite{rankingloss}.
Note that, though these prompts are learned from text descriptions solely, they can be readily deployed to classify whole images as well as image patches during testing (see Fig.~\ref{fig:intro}(c)). 
Experimental results show that, without using any labeled images, our TaI prompting surpasses zero-shot CLIP~\cite{clip} by a large margin on multiple benchmarks, \eg, MS-COCO, VOC2007, and NUS-WIDE. 




Moreover, when images are also available during training, our TaI prompting can be combined with existing methods of prompting from images to improve its performance. 
In particular, given a few annotated images, our TaI-DPT can be integrated with CoOp as a prompt ensemble for improving classification accuracy.
With partially labeled training data being provided, we may also combine TaI-DPT and DualCoOp~\cite{dualcoop} to improve multi-label recognition accuracy consistently.
Extensive results verify the effectiveness of our TaI-DPT, using in isolation or in combination, in comparison to state-of-the-arts.
%
%

To sum up, the contribution of this work include:
\begin{itemize}
\setlength{\itemsep}{0pt}
\setlength{\parsep}{0pt}
\setlength{\parskip}{0pt}
\item We propose Texts as Images in prompt tuning (\ie, TaI prompting) to adapt VL pre-trained models to multi-label image recognition.
Text descriptions are easily accessible and, in contrast to images, their class labels can be directly derived, making our TaI prompting very compelling in practice.
%
\item We present double-grained prompt tuning (\ie TaI-DPT) to extract both coarse-grained and fine-grained embeddings
for enhancing multi-label image recognition. %
Experiments on multiple benchmarks show that TaI-DPT achieves comparable multi-label recognition accuracy against state-of-the-arts.
\item  The prompts learned by TaI-DPT can be easily combined with existing methods of prompting from images in an off-the-shelf manner, further improving multi-label recognition performance.
\end{itemize}

\begin{figure*}
  \centering
     \includegraphics[width=1.0\linewidth]{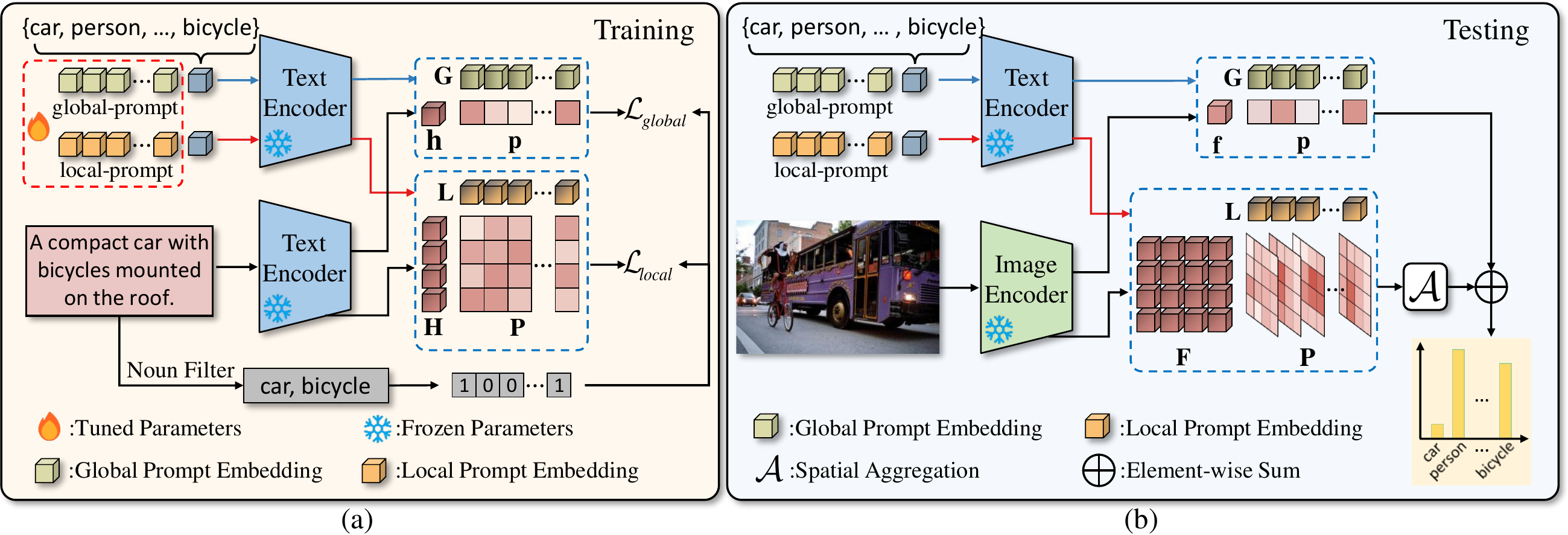}
  \vspace{-7.5mm}
  \caption{Training and testing pipeline of our proposed Text-as-Image (TaI) prompting, where we use text descriptions instead of labeled images to train the prompts. (a) During training, we use two identical text encoders from pre-trained CLIP to extract the global \& local class embeddings ($\mathbf{G} \& \mathbf{L}$) and overall \& sequential text embeddings ($\mathbf{h} \& \mathbf{H}$) respectively from the prompts and text description. The corresponding cosine similarity ($\mathbf{p} \& \mathbf{P}$) between the embeddings are guided by the derived pseudo labels with ranking loss. (b) During testing, we replace the input from text descriptions to images. The global and local class embeddings can discriminate target classes from global \& local image features ($\mathbf{f} \& \mathbf{F}$). The final classification results are obtained by merging the scores of the two branches.}
  \label{fig:pipeline} \vskip -0.1in
\end{figure*}

\section{Related Work}
\label{sec:relatedworks}

\subsection{Multi-Label Image Recognition}
Multi-label image recognition~\cite{ye2020add,benbaruch2020asymmetric,chen2019multi,chen2019multi2,guo2019multi,gao2020multi,zhang2018multilabel,he2018reinforced,wei2015hcp} aims to recognize all the object categories~\cite{coco,voc2007} or concepts~\cite{nuswide} in an input image.
%
%
To cope with multi-label images that are content-rich, various modules~\cite{wang2016cnn,chen2019learning} have been introduced to better represent the inter-class relationships and modern classification losses~\cite{benbaruch2020asymmetric,rankingloss} have been used to make model learning easier.

To model the label dependencies, CNN-RNN~\cite{wang2016cnn} introduces recurrent neural networks, \eg, RNN and LSTM, to predict appeared classes in a sequential manner. 
\cite{chen2019multi,chen2019learning,ye2020add,wang2020fast} use graph convolution modules to learn the correlation between class labels.
CHAMP~\cite{vaswani2022all} measures the severity of misclassification by building a domain-specific hierarchy tree according to the relation of categories, where each class are related to a tree node, to improve the robustness of the model.
Albeit effective, these methods requires a considerable number of labeled images to let the models learn the category relationships sufficiently. While in data-limited or label-limited regimes, \eg, few-shot or partial-label data, it will be difficult for these models to learn well as expected. 
Specifically designed loss functions also struggle to obtain significant improvements when learning with limited data.
%

\noindent \textbf{Multi-Label Recognition from Few-shot Samples.} To better exploit the small number of samples, LaSO~\cite{alfassy2019laso} synthesizes samples by manipulates the features of paired training images. 
Different ways of manipulating label sets are used to train the model, resulting in generalizable discriminative features. \cite{simon2022meta} introduces a meta-learning framework for better learning of past tasks and generalization to new tasks, and leverages the number of labels as useful information for learning.

\noindent \textbf{Multi-Label Recognition from Partial-label Data.} Partial-label refers to the scenarios where some labels are unknown. \cite{durand2019learning} propose a normalized BCE loss to balance the proportion of known labels.
\cite{chen2022structured} learns to complement unknown labels by utilizing within-image and cross-image semantic correlations.
\cite{pu2022semantic} blends the representation of training images and class proxies to compensate the loss of information due to unknown labels.

Albeit significant progress has been made, it remains a challenging issue for learning multi-label image recognition in image-limited or label-limited regimes. 
Built upon VL pre-trained models, this paper suggests to generate prompts from text descriptions instead of images, thereby offering a novel yet complementary perspective for handling low resource multi-label image recognition.


\subsection{Prompt Tuning for Vision-Language Models}

To transfer pre-trained knowledge to downstream tasks in data-limited settings, prompt tuning~\cite{coop,zhu2022prompt,vpt,cpt,nps,ge2022domain} has become a popular parameter-efficient way to achieve the goal, due to its flexibility and ease of use.
CoOp~\cite{coop} learns the prompts by using (a few) annotated images of each class from target dataset. CoCoOp~\cite{cocoop} further proposes to improve CoOp~\cite{coop} by formulating the prompts in an image-conditional way to maintain better generalization to unseen classes.
To avoid overfitting, ProGrad~\cite{zhu2022prompt} leverages predictions from zero-shot CLIP to regularize gradients in prompt learning process.
TPT~\cite{shu2022test} suggests to optimize test-time prompts by promoting the consistency of augmented test images. 
ProDA~\cite{lu2022prompt} uses multiple pieces of prompts to estimate the distribution of classifier weights for better handle of varying visual features.
DualCoOp~\cite{dualcoop} firstly adapts CLIP to multi-label image recognition with partially labeled data by learning pairs of positive and negative prompts for each class to ensure independent binary classification for each class.

Albeit existing prompt tuning approaches have achieved significant improvements in downstream tasks, images as well as a portion of class labels are prerequisite to supervise the optimization of the learnable prompts. 
In this paper, we propose to treat texts as images in prompt tuning, which, compared to labeled images, are much easier to collect with existing caption datasets and modern search engines. 
Our proposed TaI-DPT surpasses zero-shot CLIP by a large margin, and can be combined with the prompts learned by existing methods of prompting from images to further boost recognition performance.

\section{Proposed Method}
\label{sec:method}
In this section, we present our proposed Text-as-Image prompting, \ie, TaI prompting, for adapting pre-trained VL models to multi-label image recognition. 
Our TaI prompting uses only easily-accessed free-form texts as training data to learn effective prompts for downstream multi-label recognition tasks. 
To begin with, We present an overview of TaI prompting in \cref{methods0}. 
Then, we introduce our preparation of training texts in \cref{methods1}. 
We further explain the design of the double-grained prompt tuning (\ie, TaI-DPT) and the training and testing procedure in \cref{methods2}, and provide the loss function used to train the model in \cref{methods3}.
Finally, we combine TaI-DPT with the existing methods of prompting from images to improve multi-label recognition performance further.
CLIP is used to introduce our methdod.

\subsection{Overview of Our Method}
\label{methods0}
~\cref{fig:pipeline} illustrates the design of our proposed TaI-DPT framework, including the training and testing phases. 
During training, we learn prompts with only supervision from texts. Two identical copies of the text encoder ${\rm Enc_T}$ from the pre-trained CLIP are used to encode the prompts and text data, respectively. 
We introduce two sorts of trainable prompts (\ie, the global prompts and local prompts) to obtain global and regional class embeddings.
A noun filtering strategy is used to generate classification pseudo-labels for each text description, which is applied to supervise the classification scores obtained by calculating the cosine similarity of class embeddings and text features. 
Only the parameters in prompts are optimized in the training phase, while the text encoders are both kept frozen.
During testing, the class embeddings are obtained by encoding the two sets of learned prompts with the text encoder ${\rm Enc_T}$ as in training, while the other input source changes from text descriptions to test images. 
Pre-trained image encoder ${\rm Enc_I}$ from CLIP is used to extract global and dense features of each test image, then computing global and local classification scores with class embeddings generated by the global prompts and the local prompts via cosine similarity. 
The final classification result is obtained by fusing the global and local classification scores.
In the following, we explain the details of the main components of our proposed method.

\begin{figure*}
  \centering
     \includegraphics[width=1.0\linewidth]{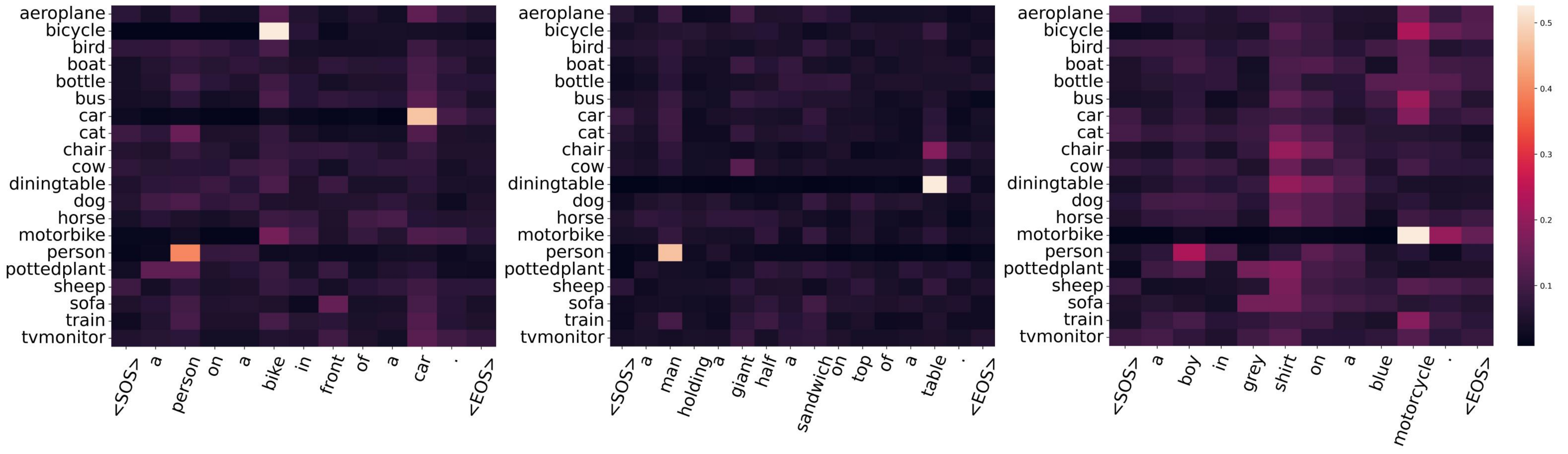}

  \caption{Visualization of correlations between the local class embedding $\boldsymbol L$ and sequential token feature from texts. 
  Each class embedding clearly correlates to words that describe the corresponding class (shown in highlight regions) rather than the global \textless EOS\textgreater~token.
  }
  \label{fig:texthm}
\end{figure*}

\begin{figure}
  \centering
     \includegraphics[width=1.0\linewidth]{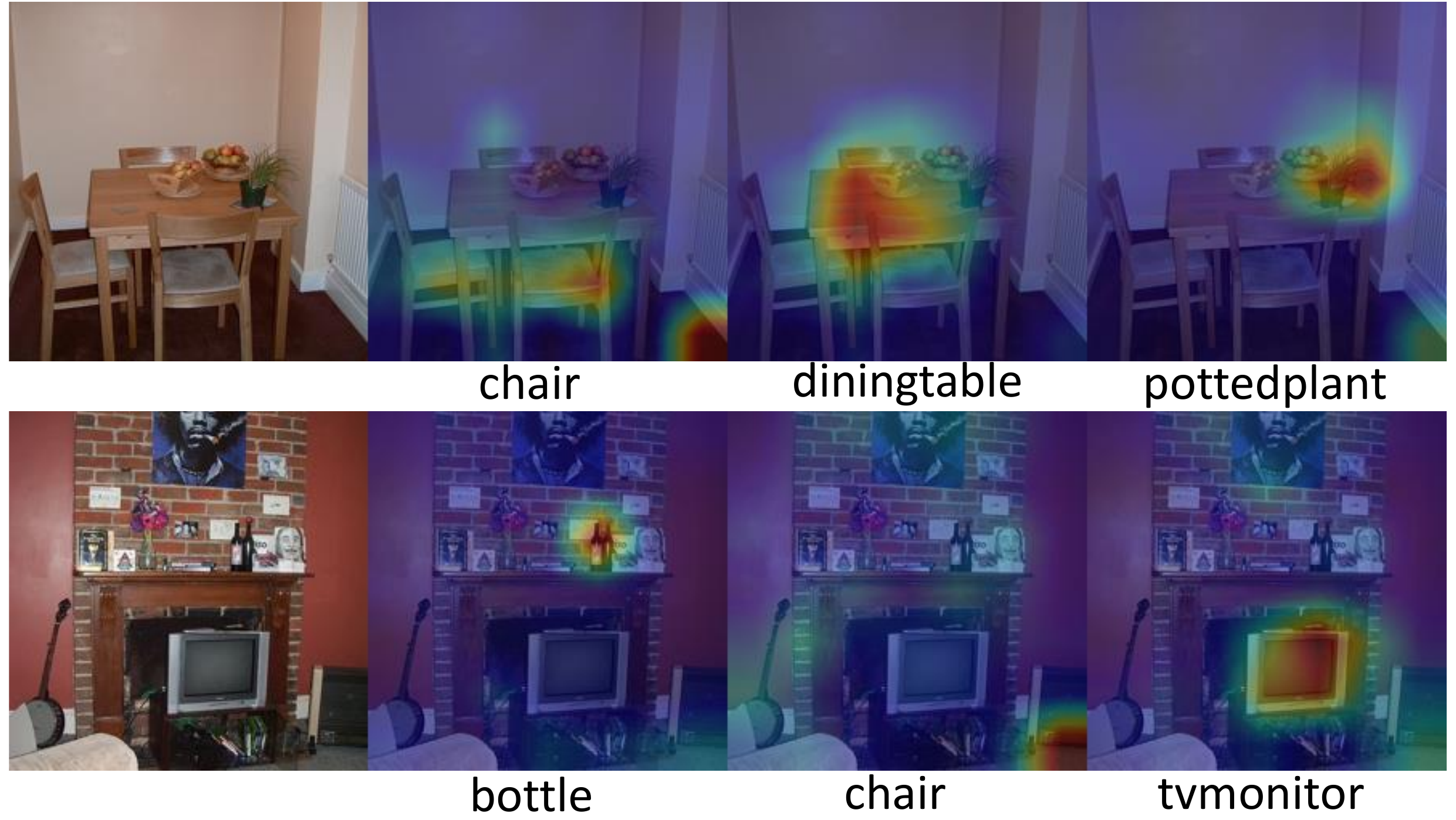}

  \caption{Visualization of correlations between the local class embedding $\boldsymbol L$ and dense image feature. The learned class embeddings can focus on the location of the object effectively.
  } \vskip -0.18in
  \label{fig:imghm}
\end{figure}

\subsection{Preparation of Text Descriptions}
\label{methods1}
To obtain sufficient category information from the language that helps in image recognition, we have to ensure that: 1) the collected text descriptions should contain rich contents that describe a relatively complete scene of an image, and 2) the contents of all text descriptions need to cover the category set of the target dataset so that the prompts can learn the discriminative features of each class well and thus obtain better recognition performance.
With an aim of ensuring reproducibility, we use captions from public image caption datasets (\emph{e.g.}, MS-COCO~\cite{coco}) and localized narratives from object detection datasets (\emph{e.g.}, OpenImages~\cite{openimages}) as our language data source, while avoiding the workloads associated with randomly crawling texts from the Internet in this paper. Note that although each caption is paired with a corresponding image and human-annotated labels, we only use the captions, and no information from the pictures and labels are disclosed during training.

For a target multi-label recognition dataset $\mathcal{X}$ that has a category set $\mathcal{S}$ = \{$\rm s_1$, $\rm s_2$, $\rm s_3$, ..., $\rm s_{\textit {C}}$\}, where \textit{C} denotes the number of categories and $\rm s_i$ denotes particular class name like ``dog'', ``plane'', etc., we search for sentences that contain at least one class name $\rm s_i$ in $\mathcal{S}$. 
%
Since multiple words or phrases usually exist to represent the same meaning for each class, searching solely for exact match of category names in texts may lead to many false negatives in the obtained pseudo ground-truth labels, which is harmful to prompt tuning.
Towards tackling this issue, we introduce a noun filter to map nouns with similar meanings into the corresponding class label. Specifically, we construct a synonym dictionary $\mathcal{D}$ by including common synonyms of each class name in the target dataset. 
If a word in a text description matches any synonym of a specific class name, it is considered to contain a description of that category. Several examples of synonyms are shown as follows:
{\small\begin{verbatim}
  {'dog','pup','puppy','doggy'}
  {'person','people','man','woman','human'}
  {'bicycle','bike','cycle'}
  {'car','taxi','automobile'}
  {'boat','raft','dinghy'}
  ...
\end{verbatim}}
\noindent More details of the synonym dictionary $\mathcal{D}$ are provided in the \emph{Suppl}.

Then we conduct noun filtration by the following steps. First, for each text description, we use the tokenizer and lemmatizer from NLTK~\cite{bird2009natural} to recover the stem of each word in the sentences. Next, for all keywords in $\mathcal{D}$, which contains all synonyms of the category set $\mathcal{S}$, we search in our language data source for sentences that contains at least one class name. For the text descriptions that do not match any synonym of any class name, we simply drop it away to ensure each piece of data has at least one concerned label. Finally, for each retained text description, we convert the class names it contains into binary pseudo-ground-truth vectors by setting classes that appear as positive and other classes as negative, following the order of class labels in the target dataset $\mathcal{X}$.

The word-level filtered labels may not be precisely correct since our searching strategy mentioned above is rather simple considering the diversity of free-form texts, where complex paraphrases and misspellings that widely exist in the corpus are not fully addressed. 
However, such a simple noun filtration can guarantee reproducibility of this work and already leads to satisfactory results of our TaI, as will be shown.
And our experiments also demonstrate that this simple and efficient data preparation lead to practical prompt tuning and compelling multi-label recognition accuracy.

\subsection{Text-as-Image for Dual-grained Prompt Tuning}
\label{methods2}

Following \cite{coop}, a prompt is defined as:
\begin{equation}
  \boldsymbol t_i = [\boldsymbol v_1, \boldsymbol v_2, \boldsymbol v_3, ..., \boldsymbol v_M, \boldsymbol s_i]
  \label{eq:0}
\end{equation}
where $i\in\{1, 2, ...,  C\} $ is the class index, $\boldsymbol s_i$ denotes word embedding of the $i$-th class name $\rm s_i$. For $j \in\{1, \ldots, M\}$, $\boldsymbol v_j$ is a learnable word embedding whose dimension is the same as the dimension of normal word embeddings in the vocabulary. 
Just like in previous methods, \eg CoOp~\cite{coop}, the prompts are learned by maximizing the probability of classifying each image into its ground-truth class:
\begin{equation}
  p(y = i| \boldsymbol x) =   \frac{\exp(\langle {\rm Enc_T}(\boldsymbol t_i),{\rm Enc_I}(\boldsymbol x)\rangle /\tau)}{{\textstyle \sum_{j=1}^{C}}\exp(\langle{\rm Enc_T}(\boldsymbol t_j), {\rm Enc_I}(\boldsymbol x)\rangle /\tau)}
  \label{eq:1}
\end{equation}
\noindent where $ \boldsymbol x $ denotes the image and $\langle \cdot, \cdot \rangle$ calculates the cosine similarity.

After large-scale pre-training with image-text contrastive loss, text features have been well-aligned to the image features of the same semantic meanings.
Therefore, based on the aligned VL representation, we advocate considering the feature of a piece of text description that describes a specific category, as an alternative to an image feature.
Given a piece of text description, optimizing the similarity between its feature representation produced by a VL model and some class embeddings is considered, for guiding the learning of prompts towards achieving categorical discriminative information.

Apart from using the global sentence representation (\ie, the coarsest-grained text feature), we find that the sequential feature of word tokens from CLIP also possesses rich fine-grained information which is very similar to the region feature of dense image feature. In CLIP~\cite{clip}, cosine similarity between global image features, obtained by visual attention pooling, and global text features, obtained by projecting the feature of the last \textless EOS\textgreater~token, are directly supervised with contrastive loss. 
In general, the global feature is sufficient for single-label classification because the target object usually is prominent in the picture. However, in multi-label recognition, the global feature is usually dominated by major objects, suppressing the recognition of non-significant objects concurrently existing in the image.
Thus, it motivates us to explore fine-grained features and avoid the domination of the overly prominent object.

%
To achieve this goal, we propose double-grained prompt tuning (\ie, TaI-DPT) that uses two sets of prompts to handle global (\ie, the coarsest-grained level) and local (\ie, the fine-grained level) features, respectively, in two parallel branches. The global prompts achieve discrimination by learning from the global feature directly learned in CLIP, while the local prompt learns from localized features. 
Formally, the double-grained prompt is defined as follows:
\begin{equation}
\begin{split}
  \boldsymbol t^G_i = [\boldsymbol v_1, \boldsymbol v_2, \boldsymbol v_3, ..., \boldsymbol v_M, \boldsymbol s_i], \\
  \boldsymbol t^L_i = [\boldsymbol v_1', \boldsymbol v_2', \boldsymbol v_3', ..., \boldsymbol v_M', \boldsymbol s_i],
  \label{eq:dpt}
\end{split}
\end{equation}
\noindent where $\boldsymbol v_j$ and $\boldsymbol v_j', j \in\{1, \ldots, M\}$ are learnable embeddings that are concatenated with word embedding $ \boldsymbol s_i$ of the $i$-th class to obtain the global prompt $\boldsymbol t^G_i$ and local prompt $\boldsymbol t^L_i$, respectively.
The sequences in Eq.~\eqref{eq:dpt} are fed to a copy of the text encoder $\rm Enc_T$ of CLIP to generate global and local class embeddings for each class, \ie $\boldsymbol G_i = {\rm Enc_T}(\boldsymbol t^G_i)$ and $\boldsymbol L_i = {\rm Enc_T}(\boldsymbol t^L_i)$,
$\boldsymbol G =\{\boldsymbol G_i\}^C_{i=1}$ and $\boldsymbol L =\{\boldsymbol L_i\}^C_{i=1}$ are encouraged to be correlated with global and local features, respectively. Note that the proposed double-grained prompts are different from dual prompts~\cite{dualcoop}, which include a pair of contrastive positive and negative prompts for each class (More discussion about the differences between our method and DualCoOp is provided in the \emph{Suppl}).

To preserve the fine-grained region features for the input image, we maintain the feature map before attention pooling layer of CLIP.
As for the input text description, we preserve the sequential token features of the entire sentence instead of only the \textless EOS\textgreater token features. 
So we have:
\begin{equation}
\begin{split}
  \{\boldsymbol f, \boldsymbol F \} = {\rm Enc_I}(\boldsymbol x), \\
  \{\boldsymbol h, \boldsymbol H \} = {\rm Enc_T}(\boldsymbol r),
  \label{eq:enct}
\end{split}
\end{equation}


\noindent where $\boldsymbol r$ denotes a piece of training text description.
$\boldsymbol f, \boldsymbol h \in \mathbb{R}^D$ are the extracted global image and text features.
$\boldsymbol F \in \mathbb{R}^{N_1\times D}$ and $\boldsymbol H \in \mathbb{R}^{N_2\times D}$ are the flattened dense image features and sequential token features, respectively, where $ N_1 = H \times W $ denotes the flattened spatial dimension of visual feature and $ N_2$ denotes the length of text tokens.

Then, the global and local similarities are computed by:
\begin{equation}
\begin{split}
  \boldsymbol p_i = \langle \boldsymbol u, \boldsymbol G_i \rangle ,
  \boldsymbol P_{ij} = \langle \boldsymbol U_j, \boldsymbol L_i \rangle
  \label{eq:cos}
\end{split}
\end{equation}
\noindent where $\boldsymbol u$ denotes either language feature $\boldsymbol h$ in training or visual feature $\boldsymbol f$ in testing, and $\boldsymbol U$ denotes $\boldsymbol H$ or $\boldsymbol F$ coordinately.
Information in local branch $\boldsymbol P$ (visualized in \cref{fig:texthm} and \cref{fig:imghm}) can be aggregated in a spatially weighted manner:
\begin{equation}
\begin{split}
  \boldsymbol p_{i}' =\textstyle \sum_{j=1}^{N} \displaystyle \frac{\exp(\boldsymbol P_{ij} / \tau_s)}{ {\textstyle \sum_{j=1}^{N}}\exp(\boldsymbol P_{ij} / \tau_s) } \cdot \boldsymbol P_{ij}
  \label{eq:spatial}
\end{split}
\end{equation}
\noindent where $\tau_s$ accommodates the extent of focusing on a specific location.
$\boldsymbol p_i$ and $\boldsymbol p_i'$ are optimized by the loss terms $\mathcal{L}_{global}$ and $\mathcal{L}_{local}$, respectively, which we will discuss in ~\cref{methods3}. And in the testing phase, $\boldsymbol p$ and $\boldsymbol p'$ are combined to obtain the final classification score.

The visualization results in \cref{fig:texthm} and \cref{fig:imghm} show that the learned local class embedding $\boldsymbol L$ can focus on each specific location where corresponding class appears, both in text descriptions and images, even if the fine-grained visual and language features are not explicitly supervised in the training of CLIP. 

\subsection{Learning Objective}
\label{methods3}
We briefly discuss the loss terms used during the training of TaI-DPT.
The overall learning objective is defined as $\mathcal{L} =\mathcal{L}_{global}+\mathcal{L}_{local}$,
where $\mathcal{L}_{global}$ and $\mathcal{L}_{local}$ are loss terms for global text embedding and local text tokens, respectively.
%
We adopt the ranking loss~\cite{rankingloss} to measure the discrepancy between classification scores and ground-truth labels, instead of a commonly used binary cross-entropy loss.
The binary cross-entropy loss is generally accompanied with a sigmoid function $\boldsymbol \sigma(x) = 1/(1 + \exp(-x)) $ to convert model outputs to probabilities. 
Nevertheless, we observe that the value of cosine similarities between image and text CLIP features $\boldsymbol p$ are not evenly distributed on either side of 0. 
Directly constraining the probability $\boldsymbol \sigma(\boldsymbol p)$ makes the optimization more difficult in this case, and this is why we employ a different ranking loss function~\cite{rankingloss}.
There may exist other options, \eg, the asymmetric loss as in ~\cite{dualcoop}.
%
%

Specifically, $\mathcal{L}_{global}$ and $\mathcal{L}_{local}$ are formulated as follows:
\begin{equation}
\begin{split}
  \mathcal{L}_{global} &\!=\! { \sum\nolimits_{i \in \{{c^+}\}}} { \sum\nolimits_{j\in \{{c^-}\}}} \max(0, m \!-\!\boldsymbol p_i \!+\! \boldsymbol p_j),\\
  \mathcal{L}_{local} &\!=\! { \sum\nolimits_{i \in \{{c^+}\}}} { \sum\nolimits_{j \in \{{c^-}\}}} \max(0, m \!-\! \boldsymbol p_i'\!+\! \boldsymbol p_j'),
  \label{eq:rl}
\end{split}
\end{equation}
\noindent where $\boldsymbol p$ and $\boldsymbol p'$ are global and aggregated local similarities described in \cref{methods2}, $m$ is the margin controlling how much higher the similarity score with the positive classes is than with the negative classes. 
%
During training, we minimize the overall objective $\mathcal{L}$ with frozen text encoders, by optimizing the global and local prompts.

\begin{figure}
  \centering
     \includegraphics[width=0.9\linewidth]{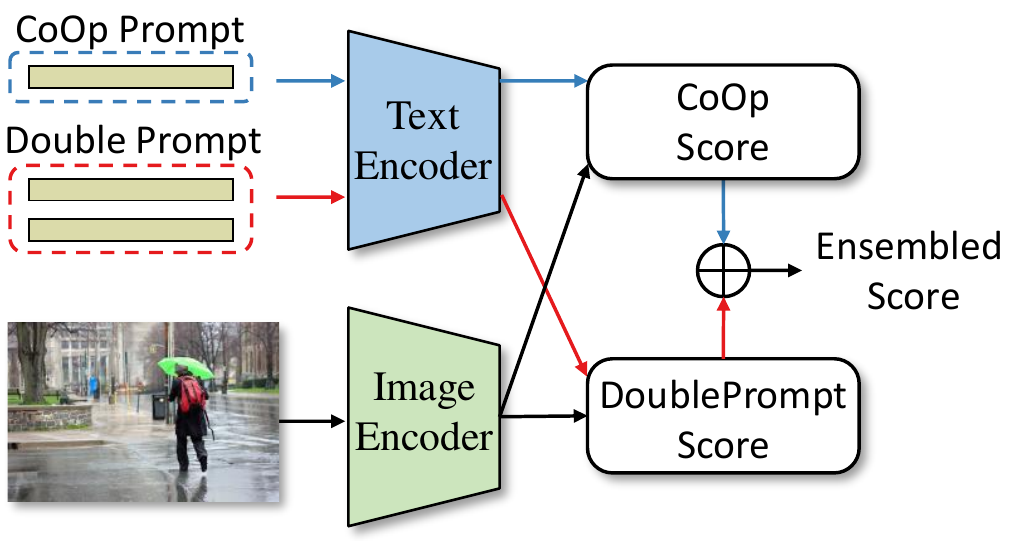}

  \caption{Our learned double-grained prompt tuning is easy to combine with existing prompt tuning methods with ensemble.}
  \label{fig:ensemble}
\end{figure}

\subsection{Incorporating with Prompting from Images}
\label{methods4}
Though our TaI-DPT is very different from existing methods of prompting from images, it is also complementary to them. 
To show this, we utilize an off-the-shelf prompt ensemble strategy to combine our TaI-DPT with existing methods in this section.
As illustrated in \cref{fig:ensemble}, using CoOp~\cite{coop} as an example, we can simply combine the scores of CoOp~\cite{coop} and that of our TaI-DPT in a weighted sum manner. 
In particular, our TaI-DPT can be integrated with CoOp~\cite{coop} when a few annotated images are provided and integrated with DualCoOp~\cite{dualcoop} when partially labeled training data are available.

We ensemble prompts by fusing the predicted scores, rather than averaging the class embeddings generated by different prompts, since the image encoder used in different methods may be different (\eg we conduct our experiments with ResNet50, while DualCoOp uses ResNet101 for partial-label prompting). So ensembling with the classification score is more convenient.
In \cref{e:3}, we also empirically show that our prompt ensemble strategy is effective in advancing multi-label recognition performance in the few-shot and partially labeled settings.

\begin{figure*}[h]
\begin{minipage}[t]{0.65\linewidth}
    \centering
    \includegraphics[width=1.0\textwidth]{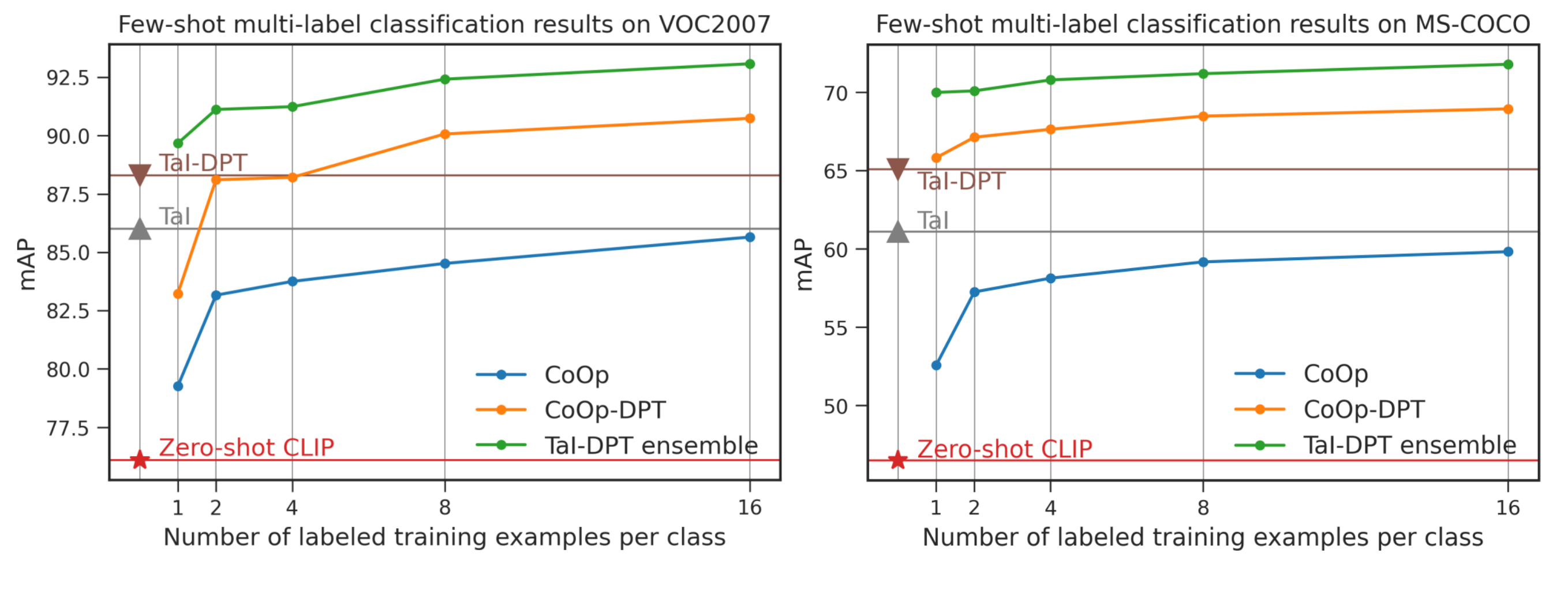}
    \vspace{-8mm}
      \caption{Comparison of different methods in few-shot multi-label recognition on VOC2007 and MS-COCO. Our zero-shot TaI-DPT can achieve comparable results with methods trained by 16-shot labeled image samples. And learned prompt ensemble proofs the complementarity between images and texts.}
    \vspace{-6mm}
    \label{fig:fewshot}
\end{minipage}
\hfill
\begin{minipage}[t]{0.32\linewidth}
\centering
    \includegraphics[width=1.0\textwidth]{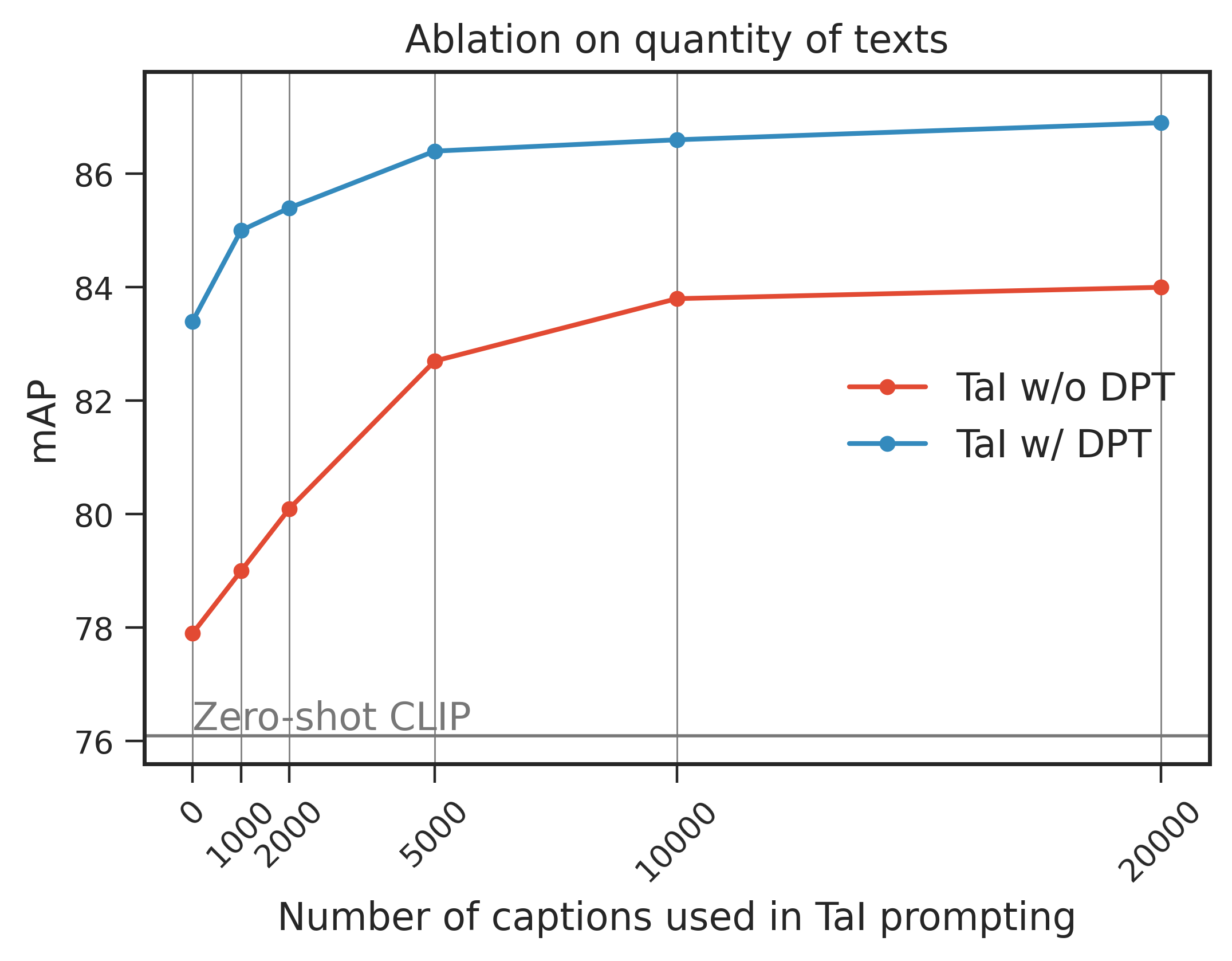} \vspace{-6mm} 
    \caption{Ablation experiment on number of texts and performance of TaI prompting on VOC2007.}
    \vspace{-6mm}
    \label{fig:ablation}
\end{minipage}

\end{figure*}

\section{Experiments}
\label{sec:experiments}

\subsection{Implementation Details}
\label{e:1}
\noindent \textbf{Architecture.} 
In our experiments, we adopt CLIP ResNet-50~\cite{clip} as the visual encoder, and use the corresponding CLIP Transformer as the text encoder.
During training, the parameters of both the two encoders are kept frozen, and only learnable prompts are optimized.

\noindent \textbf{Learnable Prompts.} 
%
Our learnable prompts are shared among classes of all datasets. 
Class-specific prompting~\cite{coop} (\ie, an individual set of parameters for each category) has also been explored, but brings limited benefits.
Hence, we adopt the shared prompts and initialize the value of each parameter with the Gaussian noise sampled from $\mathcal{N}(0, 0.02)$.
In our experiments, the length of both the global prompts and local prompts are set to $M$ = 16, while a longer sequence brings trivial improvements. 

\noindent \textbf{Datasets.} 
%
To evaluate our TaI-DPT, we conduct the experiments on VOC2007~\cite{voc2007}, MS-COCO~\cite{coco}, and NUS-WIDE~\cite{nuswide}.
VOC2007 contains 20 common categories, and following~\cite{chen2019multi,chen2019learning,dualcoop}, we form the training/test set based on the official \texttt{trainval}/\texttt{test} split (5,011 images/4,952 images).
MS-COCO includes 80 categories, and following the official split, we take 82,081 images to form the training set and 40,504 images to form the validation set. 
NUS-WIDE includes 81 concepts, which have certain inclusion relationships. We adopt its test set (107,859 images) to evaluate our method.
For zero-shot experiments in \cref{e:2}, the training sets of the datasets are not used, and we use only text data to learn the prompts as mentioned in \cref{methods1}.
Besides, for VOC2007 and MS-COCO, the language data sources are captions from MS-COCO.
%
For NUS-WIDE, we introduce localized narratives from OpenImages~\cite{openimages}, which have a broader range of content, to cover all the concepts in NUS-WIDE.
In \cref{e:3} and \cref{e:4}, for each dataset, the corresponding training data is used to conduct the experiments of partial-label and few-shot multi-label classification.

\noindent \textbf{Training Details.}
%
%
We adopt SGD optimizer to learn our prompts, and the training epochs is set to 20 for all datasets.
The learning rates for MS-COCO, VOC2007, and NUS-WIDE are empirically initialized with 1e-4, 1e-4, and 1e-3, and decay by the cosine annealing rule during training.
For ranking loss, we choose $m=1$, and scale the $\boldsymbol p$ and $\boldsymbol p'$ by a factor of 4.
%
%
$\tau_s$ is set as 0.02 via validation.

\begin{table} [!h]
    \centering
    \caption{Comparison with zero-shot methods on VOC2007, MS-COCO, and NUS-WIDE. Our proposed TaI-DPT outperforms CLIP~\cite{clip} by a large margin on all datasets.}
    \vspace{-2mm}
    \setlength{\tabcolsep}{1.8mm}
    \begin{tabular}{c|c|ccc}
    \toprule
        Method & DPT & VOC2007 & MS-COCO & NUSWIDE  \\ \hline

        \multirow{2}{*}{ZSCLIP} & \ding{55} & 76.2 & 47.3 & 36.4\\ 
        & \checkmark & 77.3 & 49.7 & 37.4 \\ \hline
        \multirow{2}{*}{TaI} & \ding{55} & 86.0 & 61.1 & 44.9  \\ 
        & \checkmark & \textbf{88.3} & \textbf{65.1} & \textbf{46.5}  \\
    \bottomrule
    \end{tabular} \vskip -0.1in
  \label{tab:zclip}
\end{table}

\begin{table*} [!ht]
  \centering
  \caption{Results of integrating our TaI-DPT with partial-label multi-label recognition method based on pre-trained CLIP. Our approach further improves the frontier performance of DualCoOp~\cite{dualcoop}. $*$ indicates the results of our reproduction.}
  \vspace{-2mm}
  \setlength{\tabcolsep}{2.3mm}
  \begin{tabular}{c|c|ccccccccc|c}
    \toprule
    Datasets & Method & 10\% &20\% & 30\% & 40\% & 50\% & 60\% & 70\% & 80\% & 90\% & Avg. \\
    \midrule
    \multirow{4}{*}{MS-COCO} & SARB~\cite{pu2022semantic} & 71.2 & 75.0 & 77.1 & 78.3 & 78.9 & 79.6 & 79.8 & 80.5 & 80.5 & 77.9\\
    & DualCoOp~\cite{dualcoop} & 78.7 & 80.9 & 81.7 & 82.0 & 82.5 & 82.7 & 82.8 & 83.0 & 83.1 & 81.9\\  
    & DualCoOp* & 81.0 & 82.3 & 82.9 & 83.4 & 83.5 & 83.9 & 84.0 & 84.1 & 84.3 & 83.3\\
    & +TaI-DPT & \textbf{81.5} & \textbf{82.6} & \textbf{83.3} & \textbf{83.7} & \textbf{83.9} & \textbf{84.0} & \textbf{84.2} & \textbf{84.4} & \textbf{84.5} & \textbf{83.6}\\ 
    
    \hline
    \multirow{4}{*}{PascalVOC 2007} & SARB~\cite{pu2022semantic} & 83.5 & 88.6 & 90.7 & 91.4 & 91.9 & 92.2 & 92.6 & 92.8 & 92.9 & 90.7 \\
    & DualCoOp~\cite{dualcoop} & 90.3 & 92.2 & 92.8 & 93.3 & 93.6 & 93.9 & 94.0 & 94.1 & 94.2 & 93.2\\  
    & DualCoOp* & 91.4 & 93.8 & 93.8 & 94.3 & 94.6 & 94.7 & 94.8 & 94.9 & 94.9 & 94.1\\  
    & +TaI-DPT & \textbf{93.3} & \textbf{94.6} & \textbf{94.8} & \textbf{94.9} & \textbf{95.1} & \textbf{95.0} & \textbf{95.1} & \textbf{95.3} & \textbf{95.5} & \textbf{94.8}\\
    
    \hline
    \multirow{2}{*}{NUS-WIDE} & DualCoOp* & 54.0 & 56.2 & 56.9 & 57.4 & 57.9 & 57.9 & 57.6 & 58.2 & 58.8 & 57.2\\  
    & +TaI-DPT & \textbf{56.4} & \textbf{57.9} & \textbf{57.8} & \textbf{58.1} & \textbf{58.5} & \textbf{58.8} & \textbf{58.6} & \textbf{59.1} & \textbf{59.4} & \textbf{58.3}\\

    \bottomrule
  \end{tabular}\vspace{-3mm}
  \label{tab:partial}
\end{table*}

\subsection{Comparison with Zero-Shot Methods}
\label{e:2}


To demonstrate the effectiveness of our proposed TaI and DPT, we first compare it with the zero-shot CLIP (ZSCLIP).
For fair comparison, we also introduce the DPT to ZSCLIP. 
Specifically, we adopt two identical default prompts ``a photo of a [CLASS]'' to separately deal with global and local features as DPT does.

\Cref{tab:zclip} lists the comparison results on VOC2007~\cite{voc2007}, MS-COCO~\cite{coco}, and NUS-WIDE\cite{nuswide} datasets.
From the table, our TaI prompting surpasses ZSCLIP by a large margin of 9.8\%, 13.8\%, and 8.5\% mAP on VOC2007, MS-COCO, and NUS-WIDE, respectively, showing the effectiveness of our TaI.
Furthermore, after training with fine-grained token features extracted from texts, our proposed DPT demonstrates a more powerful capability of discriminating local features than the default hand-crafted prompts and single global prompts.



\begin{table}[h]
    \centering
    \caption{Comparison with existing multi-label few-shot learning methods on MS-COCO. The evaluation is based on mAP for zero-shot, 1-shot and 5-shot with 16 novel classes.}
    \vspace{-2mm}
    \setlength{\tabcolsep}{3.8mm}
    \begin{tabular}{c|ccc}
    \toprule
        Method & \textbf{0-shot} & 1-shot & 5-shot \\ 
        \midrule
        LaSO~\cite{alfassy2019laso} & - & 45.3 & 58.1 \\ 
        ML-FSL~\cite{simon2022meta} & - & 54.4 & 63.6 \\ 
        TaI-DPT & 59.2 & - & - \\ 
    \bottomrule
    \end{tabular}\vspace{-5mm}
    \label{tab:laso}
\end{table}

\subsection{Comparison with Few-Shot Methods}
\label{e:3}



We further compare with multi-label few-shot learning methods to verify the effectiveness of our TaI-DPT. 
In contrast to the well-studied single-label few-shot classification problem, few works tackle the multi-label few-shot scenario. Existing methods \cite{alfassy2019laso,simon2022meta} often deploy models trained on seen classes to few-shot novel classes. In \Cref{tab:laso}, we compare our zero-shot TaI-DPT to few-shot methods on 16 novel classes (we refer readers to \cite{alfassy2019laso} for details about data split). Our TaI-DPT is comparable to the methods trained on 5-shot samples.

Besides, we consider a new multi-label few-shot setting where all the classes are regarded as novel classes. We select 1, 2, 4, 8, and 16-shot samples for each category following the strategy in \cite{alfassy2019laso}.
For fair comparison, we train CoOp~\cite{coop} and our TaI in the same settings, and we also extend them with DPT for a more comprehensive comparison. For CoOp-DPT, we set two sets of learnable prompts, to deal with global and local features, respectively.
The results are illustrated in \cref{fig:fewshot}.
One can see that, even without any image information regarding novel classes, our TaI can achieve comparable results to CoOp trained on 16-shot.
Similar trends with the MS-COCO dataset and the DPT setting support our observation that the discriminative feature of text data can be used as images for prompting.
Moreover, benefiting from the flexibility of prompts, we can easily integrate our TaI-DPT with CoOp-DPT by utilizing prompt ensembles. 
As illustrated in \cref{fig:fewshot}, though CoOp-DPT has achieved a high accuracy, combining our prompts learned with text data still brings further improvement on recognition performance.
This also proves that texts and images are complementary to each other to some extent.

\subsection{Integration with Partially Labeled Methods}
\label{e:4}

Following~\cite{dualcoop}, we conduct the experiments of multi-label recognition with partial-labeled images.
We reproduce DualCoOp on partial-labeled VOC2007 and MS-COCO with the same experimental setting as reported (reproduced results are marked with *) and explore the enhancement brought by integration with TaI-DPT. 
The results are reported in \Cref{tab:partial}. With no prior knowledge from pre-trained models, previous forefront method like SARB~\cite{pu2022semantic} struggles to learn from incomplete labels. While DualCoOp~\cite{dualcoop} achieves promising performance by prompting with images, TaI-DPT can still bring further improvements.

\subsection{Ablation Study}
\label{e:ablation}
To thoroughly investigate the effect of each component, we conduct a series of ablation studies on the quantity of texts, training loss, ensemble weight, and texts \emph{v.s.} images for prompting. 
More details are shown in the \emph{Suppl}.

\noindent\textbf{Quantity of texts. } Here, we mainly discuss the the effect of the number of text descriptions used in training on the performance of TaI-DPT on VOC2007. Following the data preparation procedure in \cref{methods1}, we end up with a total number of 66087 pieces of text that contain descriptions for 20 categories involved in VOC2007.
We test the performance of TaI-DPT with different numbers of randomly selected texts, and the results are shown in \cref{fig:ablation}. 
When no collected texts are available, 80 templates of hand-crafted prompts from \cite{clip}, like ``a cropped photo of a [CLASS]'', are used for training (all templates are shown in the \emph{Suppl}), and each template sentence correlates with one positive label corresponding to the class name inserted in [CLASS]. The increasing number of texts gradually forms a complete description of target categories, and the relationship between classes is also better characterized, which results in ascending performance.

%



\section{Conclusion}
\label{sec:conclusion}
In this paper, we propose a new view of treating texts as images in prompt tuning (\ie TaI), which learns the prompt from discriminative features of text descriptions. Compared to prior prompt tuning methods trained with images, our TaI benefits from the easy accessibility of scalable content-rich texts, which enables prompt tuning for vision tasks (\eg, multi-label image recognition) even without downstream image data. 
Double-grained prompting is further introduced to utilize both the global and fine-grained features for better multi-label recognition ability.
Nonetheless, when few-shot image samples or partial-labeled images are available, our TaI-DPT can conveniently integrate with existing prompting methods.
Experiments on MS-COCO, VOC2007, and NUS-WIDE show the validity of our proposed method.

{\small
\bibliographystyle{ieee_fullname}
\bibliography{egbib}
}

\newpage

\appendix
\section{Appendix Overview}
\label{sec:overview}
Here we provide more information of our TaI-DPT and experimental results. The appendix is organized as follows.
In \cref{sec:datasets}, we present more details about our prepared text data used for training.
In \cref{sec:ablation_supp}, we display more ablation study on the training loss, texts v.s. images for prompting and the coefficients used in the prompt ensemble.
In \cref{sec:relatedworks}, we discuss the connection and distinction between our prompt design and existing methods.

\section{More Details about Text Descriptions}
\label{sec:datasets}

\subsection{Noun Filtration}
To extract the category labels from texts exhaustively, we construct synonym dictionaries for classes involved in VOC2007~\cite{voc2007}, MS-COCO~\cite{coco}, and NUS-WIDE~\cite{nuswide} by gathering the expressions of the classes from different sources. We use the WordNet~\cite{wordnet} interface provided by \cite{bird2009natural} to get a relatively comprehensive list of synonyms and then manually select words with specific meanings for inclusion in the synonym dictionary. In addition, we also collect expressions for categories from standard online dictionaries. Besides, some words exist in the corpus in simple and compound forms, like ``cellphone'' and ``cell phone'', and we prioritize compound word matches.
Since the 80 categories of MS-COCO~\cite{coco} cover the categories of VOC2012~\cite{voc2007}, for these two datasets, we filtered the captions from MS-COCO using the same synonym dictionary (shown in ``synonyms\_COCO.txt'') to obtain the texts and labels as the training data. 
%
For NUS-WIDE~\cite{nuswide}, we introduce localized narratives from OpenImages~\cite{openimages}, which have a broader range of content, to cover all the concepts in NUS-WIDE. The synonym dictionary for NUS-WIDE is shown in ``synonyms\_NUSWIDE.txt''.


\subsection{Hand-craft Prompt Templates}
Using the noun filtration strategy above, we end up with 66,087, 100,543, and 456,759 pieces of texts for VOC2007, MS-COCO and NUS-WIDE, respectively.
Even for some common categories, the amount of texts is relatively sufficient, but we still find that there are few occurrences of certain categories in the texts. Especially for objects that are not prominent on which the text descriptions tended not to focus. So to process these categories better, we also added the hand-crafted prompt templates for each class as training data. The used templates are listed in ``prompt\_templates.txt''.
\begin{table}
    \centering
    \caption{Comparison of the results when train TaI-DPT with different learning objectives. Ranking loss (RL)~\cite{rankingloss} serves as a properer and more flexible way to guide the learning of prompts.}
    \vspace{-2mm}
    \setlength{\tabcolsep}{3mm}
    \begin{tabular}{c|ccc}
    \toprule
        Loss & VOC2007 & MS-COCO & NUSWIDE  \\ \hline
        BCE & 84.9 & 59.0 & 40.5  \\ 
        ASL\cite{benbaruch2020asymmetric} & 84.6 & 56.9 & 36.0  \\ 
        RL\cite{rankingloss} & \textbf{88.3} & \textbf{65.1} & \textbf{46.5}  \\
    \bottomrule
    \end{tabular}
  \label{tab:loss}
\end{table}

\begin{table}
    \centering
    \caption{The results of training the double-grained prompt with text data and labeled images on VOC2007. Our TaI-DPT can learn effective prompts in zero-shot setting.}
    \vspace{-2mm}
    \setlength{\tabcolsep}{3mm}
    \begin{tabular}{c|c|ccc}
    \toprule
        Method & DPT & ZSCLIP & TaI & Image \\ \hline
        \multirow{2}{*}{VOC2007} & \ding{55} & 76.2 & 86.0 & 90.0 \\ 
        & \checkmark & 77.3 & 88.3 & 93.9 \\
    \bottomrule
    \end{tabular}
  \label{tab:tai}
\end{table}

\section{More Ablation Studies}
\label{sec:ablation_supp}

\subsection{Loss Function}
\label{sec:loss}
As explained in Sec.~\textcolor{red}{3.4} of our main paper, we discussed the loss function used to train our TaI-DPT. Here, we provide the results on the three datasets when training with common binary cross-entropy loss (BCE), asymmetric loss (ASL)~\cite{benbaruch2020asymmetric} and ranking loss (RL)~\cite{rankingloss}.
Formally, the binary cross-entropy loss is defined as:
\begin{equation}
\begin{split}
  &\mathcal{L} \!=\! {\rm BCE}(\boldsymbol p, \boldsymbol y) + {\rm BCE}(\boldsymbol p', \boldsymbol y), \\
  {\rm BCE}(\boldsymbol q, \boldsymbol y) \!=\!& -\frac{1}{C} { \sum_{i \!=\! 1}^{C} } {[ \boldsymbol y_i \cdot {\rm log~}\boldsymbol q_i + (1 - \boldsymbol y_i) \cdot {\rm log~}(1-\boldsymbol q_i)]}
  \label{eq:bce}
\end{split}
\end{equation}
\noindent where $\boldsymbol p$ and $\boldsymbol p'$ are global and local classification score. And the asymmetric loss is defined as:
\begin{equation}
\begin{split}
  \mathcal{L} &\!=\! {\rm ASL}(\boldsymbol p, \boldsymbol y) + {\rm ASL}(\boldsymbol p', \boldsymbol y), \\
  {\rm ASL}(\boldsymbol q, \boldsymbol y) &\!=\! -\frac{1}{C} { \sum_{i \!=\! 1}^{C} } {[ \boldsymbol y_i \cdot k_+ + (1 - \boldsymbol y_i) \cdot k_-]},\\
  k_+ &\!=\! (1-\boldsymbol q_i)^{\gamma_+}{\rm log~}\boldsymbol q_i,\\
  k_- &\!=\! (\boldsymbol q_i^m)^{\gamma_-}{\rm log~}(1-\boldsymbol q_i^m)
  \label{eq:asl}
\end{split}
\end{equation}
\noindent where $\boldsymbol q^m = {\rm max}(\boldsymbol q - m, 0)$ and hyperparameters $\gamma_+$, $\gamma_-$ and $m$ are set as 1, 2 and 0.05, respectively, according to \cite{dualcoop}. The training results with different losses are shown in \Cref{tab:loss}.

\subsection{Texts v.s. Images for Prompting}
\label{sec:tai_supp}
To directly compare the difference between prompting with texts and prompting with images, we train our double-grained prompt with images (I-DPT) from trainval set and compare it with TaI-DPT on the test set of VOC2007~\cite{voc2007}. The results are shown in \Cref{tab:tai}. It's obvious that we can learn the prompts well with sufficient labeled images, improving the mAP of zero-shot CLIP from 77.3 to 93.9. However, when no image data is available, our TaI-DPT can reach 88.3 mAP, demonstrating the effectiveness of our zero-shot prompt tuning scheme.


\begin{figure}
  \centering
     \includegraphics[width=0.95\linewidth]{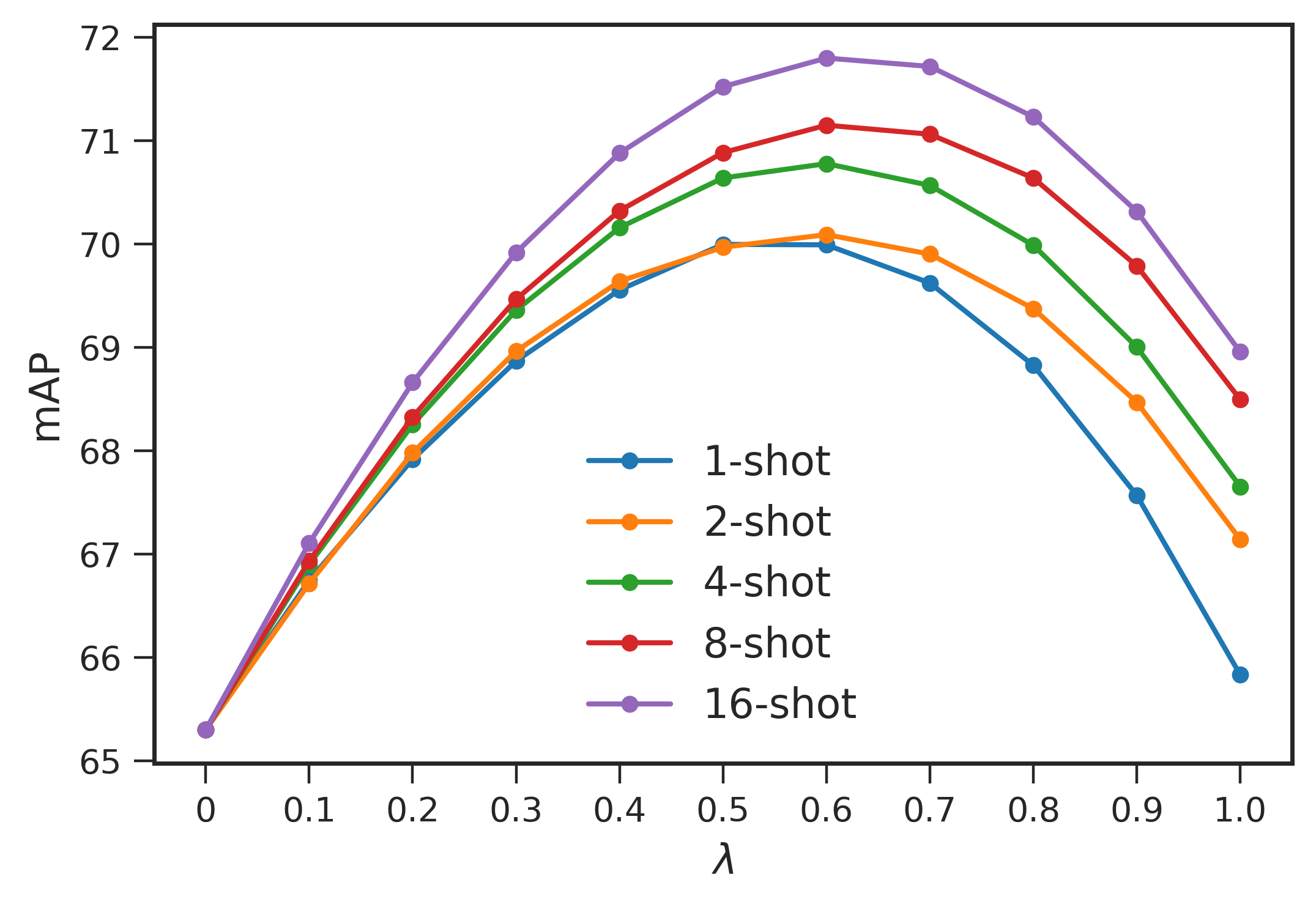}
  \vspace{-3mm}
  \caption{Relation between ensemble performance on MS-COCO and summation coefficient.}
  \label{fig:ensemble_coeffi}
\end{figure}

\subsection{Summation Coefficient in Prompt Ensemble}
\label{sec:ensemble_supp}
As illustrated in Sec. \textcolor{red}{3.5} of our main paper, our TaI-DPT can easily combine with existing prompting methods learned with images and yield complementary improvements. Here, we explore the coefficient used to fuse the classification score produced by different models. For example, let $\boldsymbol p_1$ denotes the score provided by CoOp~\cite{coop} and $\boldsymbol p_2$ denotes the score yielded by our TaI-DPT. The merged score is obtained by weighted summation $\boldsymbol p = \lambda \cdot \boldsymbol p_1 + (1 - \lambda ) \cdot \boldsymbol p_2$. 

From \cref{fig:ensemble_coeffi} we can see the change of mAP of $\boldsymbol p$ relative to coefficient $\lambda$.
So we set $\lambda = 0.6$ for the ensemble of TaI-DPT and CoOp-DPT learned from few-shot samples, which gives better results in various few-shot settings. Similarly, we set $\lambda = 0.9$ when combining our TaI-DPT with DualCoOp~\cite{dualcoop} when partially annotated images are available.

\begin{table}
    \centering
    \caption{The results of using \textbf{hand-crafted} positive and negative templates during the zero-shot inference of CLIP~\cite{clip}. Despite containing a completely opposite meaning, the negative linguistic inputs still achieve considerable accuracy.}
    \vspace{-2mm}
    \setlength{\tabcolsep}{3mm}
    \begin{tabular}{c|ccc}
    \toprule
        Template & VOC2007 & MS-COCO & NUSWIDE  \\ \hline
        Pos. & 76.2 & 47.3 & 36.4  \\ 
        Neg. & 66.2 & 41.8 & 24.3  \\ 
    \bottomrule
    \end{tabular}
  \label{tab:neg}
\end{table}

\section{Comparison with DualCoOp}
\label{sec:dualcoop}
As the first approach to adapt pre-trained CLIP~\cite{clip} to multi-label recognition tasks, DualCoOp\cite{dualcoop} proposes to use a pair of contrastive positive and negative prompts to generate binary classification probability for each class. However, the negative prompt may not be a property way to adapt CLIP. In \Cref{tab:neg} we show zero-shot recognition results of CLIP~\cite{clip} with hand-crafted positive and negative templates. We use a positive template, "a photo of a [CLASS]" and a negative template, "a photo without [CLASS]". It seems that the negative prompt is dominated by the [CLASS] token and still gives rise to considerable recognition accuracy as the positive prompt does, which can make it reluctant to analyze the effect of a negative prompt.

But for our proposed double-grained prompt tuning (DPT), the two prompts are all positive and focus on global and local features separately. Intuitively, the global prompt can be seen as a hand-crafted prompt like ``a photo of a [CLASS]'', and the local prompt can be seen as ``a cropped photo of a [CLASS]''. The two positive prompts can be learned flexibly with ranking loss~\cite{rankingloss}, without relying on each other to produce a classification score for each class. 

Besides, DualCoOp\cite{dualcoop} uses all images from the training set with partial labels to learn the prompts. Our TaI-DPT advocates using descriptive texts as an alternative when there is no image data, and the pseudo-label for each text derived with noun filtration can be regarded as incomplete categorical labels. As such, our prepared text data is somewhat homogeneous with the partial-labeled image data, which leads to gentle improvements when combining our method with DualCoOp. However, in the case of few-shot image samples available, our TaI-DPT brings considerable enhancements by ensemble with the few-shot approach like CoOp~\cite{coop} as shown in \cref{fig:ensemble_coeffi}.

%

\end{document}